\documentclass[preprint,12pt]{elsarticle}

\usepackage{amssymb}
\usepackage{textcomp}
\usepackage{makecell}
\usepackage{multirow}
\usepackage{soul, color}
\usepackage{algorithm}
\usepackage{amsbsy}
\usepackage{amsmath}
\usepackage{algpseudocode}
\usepackage{url}
\usepackage{graphicx}
\usepackage{makecell}
\usepackage{amsmath}
\usepackage{stfloats}
\usepackage{tabularx}
\usepackage{adjustbox}

\journal{Pattern Recognition}

\begin{document}
\biboptions{numbers,sort&compress} 

\begin{frontmatter}



\title{Personalized Interpretable Classification}


\author[authorLabel1]{Zengyou He}
\author[authorLabel1]{Pengju Li}
\author[authorLabel1]{Yifan Tang}
\author[authorLabel1]{Lianyu Hu}
\author[authorLabel1]{Mudi Jiang}
\author[authorLabel2]{Yan Liu}

\affiliation[authorLabel1]{organization={School of Software, Dalian University of Technology},
            country={China}}

\affiliation[authorLabel2]{organization={School of Software Engineering, Dalian University},
            country={China}}

\begin{abstract}
How to interpret a data mining model has received much attention recently, because people may distrust a black-box predictive model if they do not understand how the model works. Hence, it will be trustworthy if a model can provide transparent illustrations on how to make the decision.  Although many rule-based interpretable classification algorithms have been proposed, all these existing solutions cannot directly construct an  interpretable model to provide personalized prediction  for each individual test sample. In this paper, we make a first step towards formally introducing personalized interpretable classification as a new data mining problem to the literature. In addition to the problem formulation on this new issue, we present a greedy algorithm called PIC (\underline{P}ersonalized \underline{I}nterpretable \underline{C}lassifier) to identify a personalized rule for each individual test sample. To improve the running efficiency, a fast approximate algorithm called fPIC is presented as well. To demonstrate the necessity, feasibility and advantages of such a personalized interpretable classification method, we conduct a series of empirical studies on real data sets. The experimental results show that: (1) The new problem formulation enables us to find interesting rules for test samples that may be missed by existing non-personalized classifiers.  (2) Our algorithms can achieve the same-level predictive accuracy as those state-of-the-art (SOTA) interpretable classifiers. (3) On a real data set for predicting breast cancer metastasis, such personalized interpretable classifiers can outperform SOTA methods in terms of both accuracy and interpretability.
\end{abstract}



\begin{keyword}



Interpretable classification \sep Rule discovery \sep Personalization \sep Transductive learning

\end{keyword}

\end{frontmatter}



\section{Introduction}\label{sec:introduction}

In many real applications such as medical diagnosis and biological data analysis, interpretable machine learning models are preferred since one also likes to understand why and how the decision is made \cite{110.1145/2939672.2939874,2wang2017bayesian}. Hence, in addition to accuracy, interpretability has become one of the well-recognized goals to be achieved when developing new machine learning models and algorithms \cite{3NEURIPS2018_743394be,4NEURIPS2021_eaa32c96}. In general, a model is considered to be interpretable when it has a transparent decision process that can be understood directly by its structure and parameters, such as rules or decision trees. For classification tasks, rule-based prediction approaches are arguably the most interpretable models since they can provide illustrations on the decision process based on simple logical rules \cite{5yu2021learning,6zhang2020diverse}.  

To construct interpretable rule-based classification models, numerous research efforts have been conducted during the past decades. On one hand, existing solutions include classical sequential covering algorithms such as CN2 \cite{7clark1989cn2} and RIPPER \cite{8cohen1995fast} and classifiers based on association rules \cite{9liu1998integrating}. On the other hand, we also observe a recent upsurge of interest on developing new rule-based interpretable classification models. The interpretability of rule-based models primarily depends on rule length and the size of the rule set \cite{10ribeiro2016should}. Generally, shorter rules and smaller rule sets lead to better interpretability.

Apart from learning an interpretable classifier in which the model is assessed as a whole, there is also a need to provide illustrations for each individual prediction \cite{10ribeiro2016should}. Towards this direction, two different strategies are typically employed in the literature. The post-processing method is to learn an interpretable model locally around an individual prediction of any classifier \cite{10ribeiro2016should,11lundberg2017unified,12chen2018learning}. In contrast, the personalized transductive learning approach constructs a unique classifier for each test sample during the model creation phase \cite{13pang2011personalized,14jahid2014personalized,15zhu2017personalized}. Such a personalized classifier is desirable because real data distribution in biomedical and clinical applications could be too complicated to be represented by only one general model \cite{15zhu2017personalized}. Therefore, a personalized model may provide a more accurate prediction and can present an implicit illustration on why the prediction is made.  

Overall, a personalized and interpretable classification model should be developed to provide a unique explanation on the prediction result for each individual test sample. Unfortunately, existing solutions are insufficient to achieve this objective very well, as elaborated below. Firstly, current rule-based classification methods do provide a logical illustration on the classification process, however, those learned rules are not specially constructed for a given test sample. Secondly, the post-processing methods (e.g. \cite{10ribeiro2016should,11lundberg2017unified,12chen2018learning}) approximate each individual prediction of a third-party model, which generally cannot provide explicit rules to illustrate  the original  classification decision. Finally, personalized transductive learning approaches (e.g. \cite{13pang2011personalized,14jahid2014personalized,15zhu2017personalized}) typically adopt the support vector machine as the base classifier in which training data is reweighted to fit the test sample. As a result, the interpretability of these classification methods is inherently lacking. Unlike black-box models, rule-based models are inherently interpretable by design.

Motivated by the above observations, we intend to address the following new data mining issue: \emph{``Can we directly and efficiently construct a personalized yet interpretable classification model for each test sample?''} To be interpretable, we try to find a best-matching rule as simple as possible for the test sample. To be personalized, such a rule is obtained by searching the training data from the scratch based on the feature-value combination of test samples. However, the number of possible rules is exponential to the number of features, making it infeasible to conduct an exhaustive search. To quickly find such a personalized rule, we present a greedy algorithm that works in a breadth-first search manner. More precisely, we first check candidate rules of length  $k$ ($=1$) and then increase the rule length to $k+1$. If the best rule of length $k+1$ cannot beat the best one of length $k$, we terminate the search and return the best rule of length $k$ to classify the test sample. The goodness of each candidate rule is evaluated based on a linear combination of precision and recall. Since the above greedy algorithm can be time-consuming on larger data sets, we further propose a fast approximate algorithm. This new method employs a preprocessing step to collect frequency distributions of all length-constrained patterns across different classes. During the prediction stage, the best-matching rule is obtained from these pre-stored patterns.

To empirically demonstrate the feasibility and advantages of such a personalized interpretable classification model, we conducted extensive experiments on real data sets. The experimental results show that: (1) Our formulation can find personalized rules for test samples that may be missed by existing rule-based classification models. These customized rules can provide some interesting explanations on the class assignment for test samples. (2) Our algorithms can achieve the same-level classification accuracy as those state-of-the-art (interpretable) classification methods.

In short, the main contributions of this paper can be summarized as follows:

\begin{itemize}
  \item We make a first step towards formally introducing the problem of personalized interpretable classification. The main benefit of such personalized predictions is the capability of identifying more interpretable rules for test samples.
  \item We present efficient algorithms to identify a personalized rule for each individual test sample. To the best of our knowledge, this is the first piece of work that creates a personalized interpretable classification model in terms of logical rules.
  \item We conduct a series of experiments to demonstrate the necessity, feasibility and advantages of such a personalized  interpretable classifiers.
\end{itemize} 

The remaining parts of this paper are organized as follows: Section 2 discusses existing research efforts that are closely related to our formulation and algorithm. Section 3 presents the problem formulation on personalized interpretable classification and algorithms that can fulfill this task. Section 4 shows the experimental results and Section 5 concludes the paper.

\section{Related work}

In Section \ref{sec:2.1}, we provide a summarization on rule-based interpretable classification algorithms. In Section \ref{sec:2.2}, we present a discussion on existing personalized classification algorithms. In Section \ref{sec:2.3}, we give a brief review on learning algorithms that try to explain an individual prediction of a third-party classifier. 

\subsection{Rule-based interpretable classification}
\label{sec:2.1}

To date, there is still no consensus on a precise definition of interpretability \cite{16murdoch2019definitions}. In practice, simplicity, predictivity and stability are generally regarded as basic requirements for interpretable models \cite{17benard2021sirus}. In the context of rule-based classification models, the simplicity of model structure can be evaluated based on the number of rules, the length of each rule and the overlap among rules. The predictivity corresponds to the classification accuracy, which has been one of the long-term goals for any classifier. The stability refers to the model robustness with respect to small data perturbations \cite{17benard2021sirus}. 

To identify classification rules, heuristic sequential covering method and divide-and-conquer strategy are typically employed by early algorithms \cite{7clark1989cn2,8cohen1995fast}. However, simply being rule-based cannot fully guarantee the interpretability \cite{6zhang2020diverse} and the final rule set is not explicitly optimized with respect to the interpretability. That is, these classical rule-based classifiers mainly focus on maximizing the classification accuracy, ignoring other interpretability measures such as simplicity and stability. 

Recently, with the renewed interest on rule-based interpretable classification models, people begin to construct classifiers that explicitly optimize both accuracy and simplicity \cite{18mita2020libre,19friedman2008predictive}. These algorithms can be roughly divided into two categories. The classification algorithms in the first category \cite{110.1145/2939672.2939874,2wang2017bayesian,6zhang2020diverse} typically adopt a two-stage pipeline: rule generation and rule selection. In rule generation, an association rule mining algorithm such as FP-Growth is first employed to produce a set of candidate rules. In rule selection, a small and compact subset of rules are selected via either heuristic algorithms or the solution to a new optimization problem. Alternatively, some algorithms extract interpretable rules from random forests \cite{17benard2021sirus,26mollas2022conclusive}. The classification algorithms in the second category \cite{3NEURIPS2018_743394be,4NEURIPS2021_eaa32c96} directly learn rules from the data by formulating the rule set discovery issue as different types of optimization problems \cite{20angelino2017learning,24qiao2021learning,25ignatiev2021scalable}. 

Despite of their seeming difference with respect to problem formulation and rule discovery algorithms, all existing rule-based interpretable classification models follow an inductive learning paradigm. That is, these classifiers conduct rule learning only on the training data, which are unable to provide a potentially customized rule for each individual test sample. 

\subsection{Personalized transductive classification}
\label{sec:2.2}

In addition to the training data, the transductive learning approach employs information from the testing data as well \cite{27vapnik1999nature}. In the context of classification, a personalized transductive classification model creates a unique model for each test sample adaptively \cite{13pang2011personalized}. Hence, such a learning paradigm naturally fits the applications where the focus is on the prediction and illustration for each individual sample.

The most representative personalized transductive classification algorithm is the \emph{k} nearest neighbor (kNN) classifier \cite{28cover1967nearest}, in which the prediction model is dynamically constructed based on samples within the kNN neighborhood. Recently, the support vector machine (SVM) is customized to generate various personalized models and these models are applied to tackle various biomedical data analysis tasks \cite{13pang2011personalized,14jahid2014personalized,15zhu2017personalized}. 

Either the lazy learning approach such as kNN or the personalized SVM cannot provide an explicit explanation on the classification decision. Hence, a personalized yet interpretable classification model is still lacking in the literature. 

\subsection{The interpretation of an individual prediction}
\label{sec:2.3}

How to interpret a black-box machine model has been widely investigated recently \cite{29guidotti2018survey,30burns2020interpreting}. There are two closely related problems towards this direction: the interpretation of a learned model and the interpretation of an individual prediction. For existing solutions on the former issue, please refer to a recent review in \cite{29guidotti2018survey}. Here we only focus on the latter issue since our algorithm also seeks to provide an explanation on the prediction for each test sample. 

To date, many effective algorithms have been proposed to learn an interpretation for each individual prediction of any classifier (e.g. \cite{10ribeiro2016should,11lundberg2017unified,12chen2018learning}). Despite of their great successes, we have the following remarks. 

First of all, these methods provide an explanation after the model selection procedure. That is, they did not construct an interpretable model for each test sample during the model selection phase in an adaptive manner. As a result, we may fail to find a more appropriate interpretation due to the separation of model construction and interpretation. Moreover, these methods generally cannot provide explanations in terms of rules. For example, the model to be explained is approximated with a linear model in a local region around the test sample in \cite{10ribeiro2016should}. And the feature importance scores derived from the weights of the linear model are employed to explain the corresponding prediction. 

\section{Method}

\subsection{Problem formulation}

Let $D=\left\{(x_i,y_i) \mid i=1,...,N \right\}$ denote the training set of $N$ samples, where $x_i$ comprises $M$ categorical feature values and $y_i\in Y$ is a class label. We use $f_j$ to denote the $j$th feature and a predicate takes the form of $f_{j} =x_{ij}$, where $x_{ij}$ is one of the possible feature values for the $j$th feature. An itemset $s$ is defined as a conjunction of  $k$ $(1 \leqslant  k \leqslant  M)$ predicates. For a given sample $(x_{i},y_{i})$, it satisfies an itemset $s$ only if all predicates in $s$ are true in the sample. We use $s \subseteq  x_{i}$ to denote the fact that $x_{i}$ satisfies $s$. A rule $r$ is a tuple $(s,y)$ where $s$ is an itemset and $y$ is a class label.

 For each test sample $(x,?)$ in which the class label is unknown, we try to find a ``best" rule $\hat{r}=(\hat{s},\hat{y})$ such that $x$ satisfies $\hat{s}$, and then $\hat{y}$ is the label we predict. Hence, the personalized interpretable classification problem can be cast as the following algorithmic issue:

\begin{itemize}
  \item Input: A training dataset $D$, a test sample $x$.
  \item Output: The ``best" rule $\hat{r}=(\hat{s},\hat{y})$ that matches $x$.
\end{itemize}

\begin{table}[htb]
  \caption{An example dataset. }
  \centering
  \renewcommand\arraystretch{1.2}
  \begin{tabular}{ccccc}
    \Xhline{1.2pt}
     \multirow{2}{*}{Class} & \multicolumn{4}{c}{Features} 
    \\
     & $f_1$ & $f_2$ & $f_3$ & $f_4$ 
     \\ \hline
      1 & $a_1$ & $b_1$ & $c_1$ & $d_1$
      \\
      1 & $a_1$ & $b_2$ & $c_1$ & $d_2$ 
      \\
      1 & $a_2$ & $b_3$ & $c_2$ & $d_1$ 
      \\
      2 & $a_1$ & $b_2$ & $c_2$ & $d_1$
      \\ 
      2 & $a_2$ & $b_3$ & $c_1$ & $d_2$ 
      \\
      2 & $a_3$ & $b_1$ & $c_2$ & $d_1$
      \\
      2 & $a_1$ & $b_2$ & $c_2$ & $d_2$
      \\
      ? & $a_1$ & $b_3$ & $c_2$ & $d_1$
    \\ \Xhline{1.2pt}
  \end{tabular}
  \label{tbl:example}
\end{table}

To illustrate above concepts, the dataset in Table~\ref{tbl:example} is taken as an example. In Table~\ref{tbl:example}, rows stand for samples and columns represent their class label and features. There are 7 training samples in which 3 samples are drawn from class 1 and another 4 samples are obtained from class 2. We take the sample in the last line as a test sample. Assume that we find a ``best" rule $\hat{r}=(f_1=a_1 \land  f_3=c_2,2)$. Obviously, the test sample satisfies this rule and it  will be classified to class $2$.

The objective of identifying the ``personalized" rule is to maximize the discriminative ability for purpose of both classification and description, subject to the condition that the length of the rule does not exceed $maxL$. For each test sample, the goal is to find the shortest possible rule with the strongest discriminative ability that can cover the sample.

Since our algorithm can only handle categorical features, we adopt a pre-processing procedure to discretize numeric features into categorical ones. More precisely, we employ the equal width discretization method to split the $j$th numeric feature values into $g_j$ groups, where $g_j$ is a user-specified parameter. 
  
\subsection{Goodness of a rule}

Obviously, the ``best" rule we find should satisfy the test sample. Apart from this necessary condition, we evaluate each candidate rule by its accuracy and simplicity. For accuracy, we use a linear combination of precision and recall. For simplicity, since we  only choose one rule so that the number of rules and the overlap among rules make no sense in our problem. Therefore, the length of a rule is the only indicator for simplicity and  shorter rules are preferred than longer ones.

For a rule $r=(s,y)$, its length $length(r)$ is defined as the  number of  predicates in $s$. The precision is defined as the ratio between the number of samples satisfying $s$ in class $y$ and the number of all samples satisfying $s$. The recall is the ratio between the number of samples satisfying $s$ in class $y$ and the number of all samples in class $y$. 
\begin{equation}
precision(r,D)=\frac{ \mid \left\{(x_{i},y_{i}) \mid y_{i}=y, s \subseteq x_{i}\right\} \mid }{ \mid \left\{(x_{i},y_{i}) \mid s \subseteq x_{i}\right\} \mid }.
\end{equation}
\begin{equation}
recall(r,D)=\frac{ \mid \left\{(x_{i},y_{i}) \mid y_{i}=y, s \subseteq x_{i}\right\} \mid }{ \mid \left\{(x_{i},y_{i}) \mid y_{i}=y\right\} \mid }.
\end{equation}

To make a trade-off between precision and recall, we use the linear combination of precision and recall:
\begin{equation}
    \label{eq:A}
    A(r,D)=\alpha * precison(r,D) + (1-\alpha )*recall(r,D),
\end{equation}
 where $\alpha$ is a user-specified parameter. 
Then, our two objectives can be summarized as follows:
\begin{itemize}
  \item For simplicity: minimize $length(\hat{r})$.
  \item For accuracy: maximize $A(\hat{r},D)$.
\end{itemize}

For example, suppose that there are two rules $r_1=(f_1=a_1 \land  f_3=c_2,2)$ and $r_2=(f_2=b_3 \land  f_3=c_2,1)$. In the training samples in Table~\ref{tbl:example}, two samples from class 2 satisfy $f_1=a_1 \land  f_3=c_2$ and one sample from class 1 satisfies $f_2=b_3 \land  f_3=c_2$. The precision of $r_1$ is the ratio between the number of samples satisfying $r_1$ in class 2 and the number of all samples satisfying $r_1$, which is $2 \div 2 =1$. The recall of $r_1$ in class 2 is calculated as the ratio between the number of samples satisfying $r_1$ in class 2 and the number of all samples in class 2, which is $2 \div 4 =0.5$. The precision and recall of $r_2$ can be calculated in the same way, whose values are  $1$ and $0.333$. If we set $\alpha=0.5$, then $A(r_{1},D)$ is  $0.75$ and $A(r_{2},D)$ is $0.667$. Hence, $r_1$ is better than $r_2$ because it has a higher accuracy  score and both rules have the same length of 2. 

\subsection{A greedy algorithm}

\subsubsection{An overview }
 
If we employ an exhaustive method for finding the ``best" rule,  the number of candidate  rules is $2^M$, where $M$ is the number of features.  Apparently, such a naive algorithm is quite time-consuming in practice. Hence, we present a greedy algorithm to find a local optimal solution in a breadth-first manner. Our algorithm identifies personalized rule for a specific input sample, providing transparent decision-making process and explanation for prediction result, making it locally interpretable. 

\begin{algorithm}[H]
    \caption{The naive  greedy algorithm.}
    \label{alg1}
    \begin{algorithmic}[1]
    	\linespread{1.2}\selectfont
      \Require A training dataset $D$, a test sample $x$, a maximal length parameter $maxL$ and the parameter $\alpha$. 
      \Ensure The ``best" rule $\hat{r}=(\hat{s},\hat{y})$ that satisfies  $x$.
      \For{$k=1 ~ $\textbf{to}$ ~ maxL $}
      \State evaluate all rules of length $k$ 
      \State $\hat{r}_k \gets $ the ``best" rule of length $k$
      \If{$A(\hat{r}_{k},D) \leqslant A(\hat{r}_{k-1},D)$}
      \State \Return $\hat{r}_{k-1}$
      \EndIf
      \EndFor
      \State \Return $\hat{r}_{k}$
  \end{algorithmic}
\end{algorithm}

As shown in Algorithm~\ref{alg1}, we start from the rules of length $k=1$ to obtain  the ``best" rule $\hat{r}_{k}$.  When $A(\hat{r}_{k},D) \leqslant A(\hat{r}_{k-1},D)$, the algorithm will be terminated  and $\hat{r}_{k-1}$ will be returned to classify the test sample. Otherwise, we will continue to examine rules of length $k+1$ until the length exceeds the maximal length parameter $maxL$. To generate candidate rules of length $k$, we employ a method that is used in the Apriori algorithm~\cite{agrawal1994fast}. That is, the rules of length $k-1$ are joined to generate candidate rules of length $k$. Note that such an operation has been widely used in the field of frequent pattern mining.

\begin{table}[htb]
  \caption{Rules of length $1$.}
  \centering
  \renewcommand\arraystretch{1.2}
  \begin{tabular*}{0.3\textwidth}{@{\extracolsep\fill}ccc}
    \Xhline{1.2pt}
      \multicolumn{2}{c}{Rule}  & $A(r,D)$
     \\ \hline
     $f_1=a_1$ & 1 & 0.583
     \\
     $f_2=b_3$ & 1 & 0.417
     \\
     $f_3=c_2$ & 2 & $\pmb{0.750}$
     \\
     $f_4=d_1$ & 1 & 0.583
    \\ \Xhline{1.2pt}
  \end{tabular*}
  \label{tbl:length1}
\end{table}

\begin{table}[htb]
  \caption{Rules of length $2$.}
  \centering
  \renewcommand\arraystretch{1.2}
  \begin{tabular}{ccc}
    \Xhline{1.2pt}
      \multicolumn{2}{c}{Rule}  & $A(r,D)$
     \\ \hline
     $f_1=a_1 \land  f_2=b_3$ & - & -
     \\
    $ f_1=a_1 \land  f_3=c_2 $ &  2 &  $\pmb{0.750}$ 
    \\
    $ f_1=a_1 \land  f_4=d_1 $ &  1 & 0.417 
    \\
    $ f_2=b_3 \land  f_3=c_2 $ &  1 & 0.667 
    \\
    $ f_2=b_3 \land  f_4=d_1 $ & 1  & 0.667 
    \\
    $ f_3=c_2 \land  f_4=d_1 $ &  2 &  0.583
    \\ \Xhline{1.2pt}
  \end{tabular}
  \label{tbl:length2}
\end{table}

To show how the algorithm works in practice, let us still utilize the data in Table~\ref{tbl:example} as an example. Firstly, we construct four candidate rules of  length $1$ from the test sample. Then we determine their class label by calculating $A(r,D)$ scores ($\alpha=0.5$) in each class. For example, $f_1=a_1$ has an accuracy score  of $0.583$ when it belongs to class $1$. Similarly, this score will be $0.5$ when it belongs to class $2$. We choose the class associated with a larger score so that  the rule is $(f_1=a_1,1)$ with a  score of $0.583$. All rules of length $1$ are listed in Table~\ref{tbl:length1}. In  a similar manner,  all rules of length $2$ are listed in Table~\ref{tbl:length2}. The ``best" rule in Table~\ref{tbl:length1} is $\hat{r}_1=(f_3=c_2,2)$ and the ``best" rule in Table~\ref{tbl:length2} is $\hat{r}_2=(f_1=a_1 \land  f_3=c_2,2)$. Since  $A(\hat{r}_{2},D) \leqslant A(\hat{r}_1,D)$, the algorithm is terminated and the test sample is classified to class 2. $\hat{r}_1$ is better than $\hat{r}_2$ because $\hat{r}_1$ is shorter when their scores are completely equal.

\subsubsection{Pruning based on the upper bound of $A(r,D)$}

To further prune the  search space, we first provide a direct  upper bound on $A(r,D)$. For precision, it has a loose upper bound of $1$. Hence, we have
\begin{equation}
A\left( r,D \right) \leqslant \alpha +\left( 1-\alpha \right) *recall(r,D).
\end{equation}
To deduce the upper bound of recall, we first define the support of the itemset $s$ of a rule $r$ with respect to a class $c$ as follows:
\begin{equation}
support(s,c)=\frac{ \mid \left\{(x_{i},y_{i}) \mid y_{i}=c, s \subseteq x_{i}\right\} \mid }{ \mid \left\{(x_{i},y_{i}) \mid y_{i}=c\right\} \mid },
\end{equation}
where $c$ is one class label in $Y$. According to definition of recall, we have
\begin{equation}
\begin{aligned}
recall(r,D)&=support(s,y)
\\
&\leqslant \underset{c\in Y}{\max}\left\{ support(s,c) \right\},
\end{aligned}
\end{equation}
where $y$ is the corresponding class label in the rule $r$. Moreover, the set of all sub-rules of $r=(s,y)$ of length $length(r)-1$ is:
\begin{equation}
R'=\left\{ r'=\left( s',y' \right)  \mid s'\subseteq  s,length\left( r' \right) =length\left( r \right) -1 \right\}.
\end{equation}
For a fixed class label $c$, we have the following inequality according to the anti-monotonicity of support: 
\begin{equation}
support\left( s,c \right) \leqslant \underset{r'\in R'}{\min}\left\{ support\left( s',c\right) \right\}.
\end{equation}
Hence we have an upper bound of $recall(r,D)$ and an upper bound $ub(r,D)$ of $A(r,D)$:
\begin{equation}
\begin{aligned}
recall(r,D)&=support(s,y)
\\
&\leqslant \underset{c\in Y}{\max}\left\{ support(s,c\right\} 
\\
&\leqslant \underset{c\in Y}{\max}\left\{ \underset{r'\in R'}{\min}\left\{ support\left( s',c \right) \right\} \right\},
\end{aligned}
\end{equation}
and 
\begin{equation}
\begin{aligned}
A\left( r,D \right) &\leqslant ub\left( r,D \right) 
\\
&=\alpha +\left( 1-\alpha \right) *\underset{c\in Y}{\max}\left\{ \underset{r'\in R'}{\min}\left\{ support\left( s',c\right) \right\} \right\}.
\label{ubdefination}
\end{aligned}
\end{equation}

\begin{algorithm}[t]
    \caption{The PIC algorithm.}
    \label{alg2}
    \begin{algorithmic}[1]
    \linespread{1.2}\selectfont
      \Require A training dataset $D$, a test sample $x$, a maximal length parameter $maxL$ and the parameter $\alpha$. 
      \Ensure The ``best" rule $\hat{r}=(\hat{s},\hat{y})$ that satisfies  $x$.
      \For{$k=1 ~ $\textbf{to}$ ~ maxL $}
      \State  $R_k \gets createCandidateRules(R_{k-1},k)$
      \For{$r~ in ~R_k$}
      \If{$ub(r,D) \leqslant A(\hat{r},D)$}
      \State remove $r$ from $R_k$
      \EndIf
      \State evaluate $A(r,D)$ and $con(r,D)$
      \State update $\hat{r}$ and $\hat{r}_{k}$
      \If{$con(r,D) \leqslant A(\hat{r},D)$}
      \State remove $r$ from $R_k$
      \EndIf
      \EndFor
      \If{$A(\hat{r}_{k},D) \leqslant A(\hat{r}_{k-1},D)$}
      \State \Return $\hat{r}_{k-1}$
      \EndIf
      \EndFor
      \State \Return $\hat{r}_{k}$
  \end{algorithmic}
\end{algorithm}

In the rule  search process, we have three different  kinds of pruning strategies. First of all,   we can evaluate $ub(r,D)$ for each rule $r$ before scanning the  training set to calculate  its $A(r,D)$. If $ub(r,D)$ is no greater than $A(\hat{r},D)$, which is the score of the best rule $\hat{r}$ found so far, we can delete rule $r$ from the candidate rule  set. Secondly,  when we generate candidate rules of length $k+1$, we only consider those rules whose all $k$ sub-rules  belong to the candidate rule set  of length $k$. Thirdly, for any rule $r''$ that contains $r$ as a sub-rule, it is easy to know that $A(r'',D) \leqslant con(r,D)$, where $con(r,D)$ is defined as below:
\begin{equation}
con\left( r,D \right) =\alpha +\left( 1-\alpha \right) *\underset{c\in Y}{\max}\left\{ support\left( s,c \right) \right\}.
\end{equation}
By evaluating $con(r,D)$, we can know the upper bound of all super-rules  of $r$. If $con(r,D) \leqslant A(\hat{r},D)$, we can remove $r$ from the candidate rule set.

The improved greedy algorithm that is equipped with the upper bound-based pruning is shown in Algorithm~\ref{alg2}. The three pruning strategies are respectively employed in line 2 (strategy two), lines 4-6 (strategy one), lines 9-11 (strategy three). In line 7, we scan the  training set to evaluate $A(r,D)$ and $con(r,D)$.

\begin{figure}[t]
  \centering
  \includegraphics[width=\columnwidth]{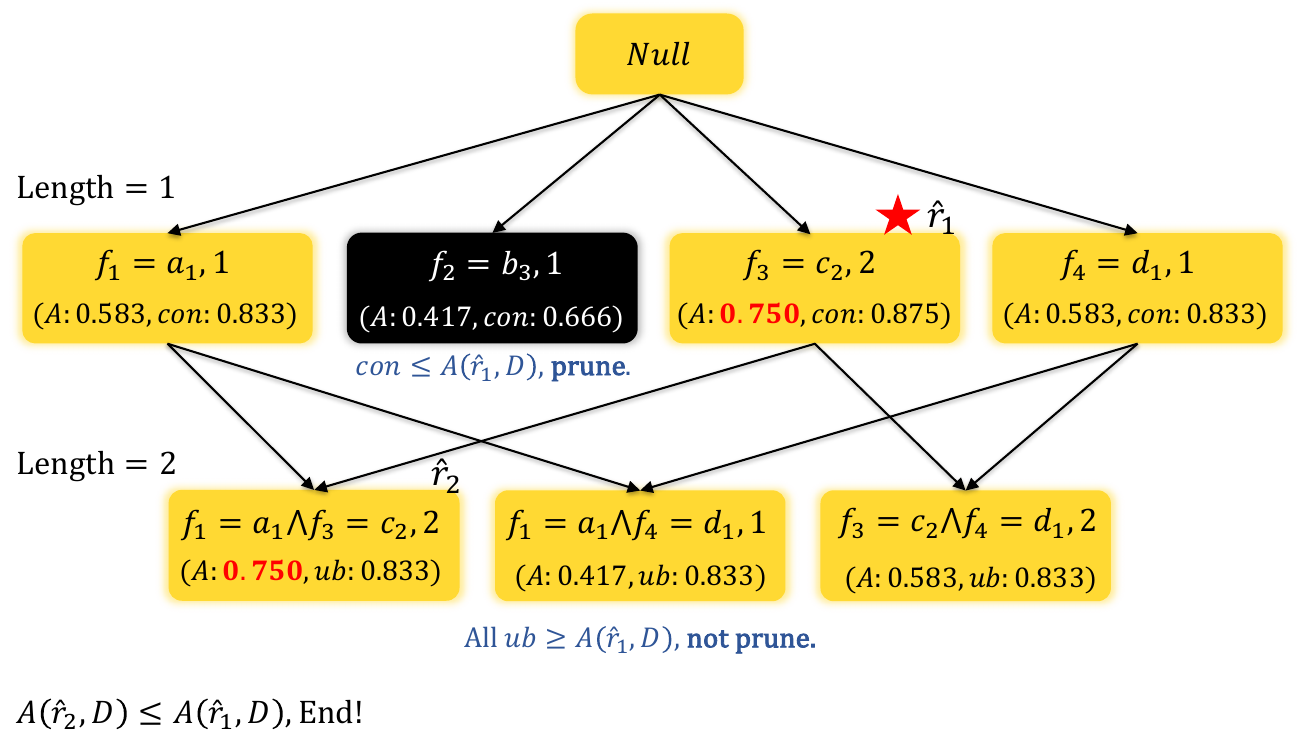}
  \caption{An illustration on the breath-first search procedure and the pruning strategy utilized by the greedy algorithm.}
  \label{fig:pic_bfs}
\end{figure}
\begin{table}[h]
  \caption{Rules of length $1$ (with pruning). }
  \centering

  \renewcommand\arraystretch{1.2}
  \begin{tabular}{cccc}
    \Xhline{1.2pt}
      \multicolumn{2}{c}{Rule}  & $A(r,D)$ & $con(r,D)$
     \\ \hline
     $f_1=a_1$ & 1 & 0.583 & 0.833
     \\
     $f_2=b_3$ & 1 & 0.417 & 0.666
     \\
     $f_3=c_2$ & 2 & $\pmb{0.750}$ & 0.875
     \\
     $f_4=d_1$ & 1 & 0.583 & 0.833
    \\ \Xhline{1.2pt}
  \end{tabular}
  \label{tbl:length1wp}
\end{table}
\begin{table}[htb]
  \caption{Rules of length $2$ (with pruning). }
  \centering
  \renewcommand\arraystretch{1.2}
  \begin{tabular}{cccc}
    \Xhline{1.2pt}
      \multicolumn{2}{c}{Rule}  & $ub(r,D)$ & $A(r,D)$
     \\ \hline
    $ f_1=a_1 \land  f_3=c_2 $ & 2 &  0.833&  $\pmb{0.750}$ 
    \\
    $ f_1=a_1 \land  f_4=d_1 $ & 1 &  0.833& 0.417 
    \\
    $ f_3=c_2 \land  f_4=d_1 $  & 2 & 0.833 &  0.583
    \\ \Xhline{1.2pt}
  \end{tabular}
  \label{tbl:length2wp}
\end{table}

Then we show how these strategies are employed in the searching progress for finding rules from the data in  Table~\ref{tbl:example}, and this process is illustrated in Figure~\ref{fig:pic_bfs}. For rules of length $1$, they do not have nonempty sub-rules and it is unnecessary to evaluate their $ub(r,D)$. All their accuracy  scores  and their $con(r,D)$ values are shown in Table~\ref{tbl:length1wp}. We find that $con(r=(f_2=b_3,1),D)=0.666 < A(\hat{r}_1,D)=0.750$, which means all super-rules  of $r=(f_2=b_3,1)$ have no opportunity to achieve a higher score than $0.666$. So this rule is removed and it will not be used for generating $R_{2}$, as shown in the black rectangle of Figure~\ref{fig:pic_bfs}. There are still  three candidate rules of length $2$ in Table~\ref{tbl:length2wp}. We evaluate their $ub(r,D)$ values and find that they are $0.833$, which are higher  than the best score $0.750$ found  so far. So we scan  the training set to calculate the  $A(r,D)$ score for each rule. Then, the algorithm is terminated because $A(\hat{r}_{2},D) \leqslant A(\hat{r}_1,D) $ and $\hat{r}_1=(f_3=c_2 , 2)$ will be returned as the identified ``best" rule, as shown at the bottom of Figure 1.

\subsection{A fast algorithm}
The Algorithm~\ref{alg2} employs a greedy strategy to mine local optimal solution. The discovery best-matching rule for predicting the class label of each test sample requires multiple iterations over the training dataset. Although a pruning technique is applied, the process remains computationally expensive. To improve the running efficiency, we introduce a semi-lazy strategy which pre-computes the frequency of each candidate itemset across different classes in the training dataset and records such information in a hash table, thereby accelerating rule evaluation and selection during the prediction stage. 

We now describe this fast personalized interpretable classification approach in Algorithm~\ref{alg3}, denoted as fPIC. The fPIC algorithm consists of two stages: the first stage processes the training data in $D$ to compute itemset frequencies, while the second stage predicts the class label of a test sample. In the first stage, we enumerate all possible itemsets of size up to $maxL$ for each training sample $(x_i,y_i)$ in $D$ (line 3). After iterating through all training samples, the frequency of each possible itemset generated from $D$ across different classes is recorded in a hash table (lines 2 to 7). The construction of hash table in the preprocessing stage is performed only once and can be used for all test samples.

\begin{algorithm}[t]
    \caption{The fPIC algorithm.}
    \label{alg3}
    \begin{algorithmic}[1]
    \linespread{1.2}\selectfont
        \Require A training dataset $D$, label set $Y$, a test sample $x$, a maximal length parameter $maxL$ and the parameter $\alpha$. 
        \Ensure The ``best" rule $\hat{r}=(\hat{s},\hat{y})$ that satisfies  
      $x$.
      
        \Statex\hspace{-1.5em}  \textbf{Preprocessing Stage:}
        \State Initialize hash table $HT$
        \For{$(x_i,y_i)~ in ~D$}
            \State $S \gets getAllComb(x_i,maxL)$ 
            \For{$s~ in ~S$}
                \State $HT[s][y_i]\mathrel{+}= 1$
            \EndFor
        \EndFor
        \Statex 
        \Statex\hspace{-1.5em} \textbf{Prediction Stage:}
        \State $S_t \gets getAllComb(x,maxL)$
        \State $R \gets S_t \times Y$
        \For{$r~ in ~R$}
            \State evaluate $A(r,D)$ according to Equation~\eqref{eq:A}
            \State update $\hat{r}$  
        \EndFor
        \State \Return $\hat{r}$
  \end{algorithmic}
\end{algorithm}
In the prediction stage, it is worth noting that, due to the use of hash table, fPIC no longer requires complex pruning strategies. Instead, it directly evaluates the scores of all possible rules for a test sample $x$ and selects the best one. Mine all itemsets  $S_t$  in  $x$  using the same method described above (line 8), and then combine $S_t$ with the class label set  $Y$  to construct a candidate rule set  $R$  for  $x$ (line 9). Ultimately, evaluate each rule in $R$ based on Equation~\eqref{eq:A} to determine the best rule $\hat{r}$ for classifying $x$ (line 10 to 13).

However, the preprocessing stage of fPIC stores the hash table that records the itemset frequencies in memory, which requires more memory compared to PIC, trading memory usage for improved algorithm running time efficiency. When the number of distinct feature values or $maxL$ is large enough, memory shortage may occur, making it difficult to continue running. In contrast, PIC does not require preprocessing and directly iterates through the training dataset for each test sample, avoiding this issue.

\section{Experiment}

In order to assess the performance of our algorithm, a series of experiments are conducted. First, we compare our algorithm with existing interpretable classification methods in terms of the predictive performance. Second, to verify the personalization and  interpretability of our method, we compare the set of all rules found by our method with the rule set reported by existing algorithms. Finally, we employ the PIC algorithm on a real dataset to demonstrate the effectiveness and rationale of our formulation.  The PIC algorithm is implemented in C++, fPIC algorithm is implemented in python and the experiments are conducted on a workstation with an Intel(R) Core(TM) CPU(11400F @ 2.60GHz) and 16GB memory.
\begin{table}[t]
  \caption{Some important characteristics of the data sets used in the experiment. $N$ represents the number of samples, $M$ is the number of features, $C$ is the number of classes.  }
  \centering
  \renewcommand\arraystretch{1.2}

  \begin{tabular}{ccccc}
    \Xhline{1.2pt}
     Dataset & $N$ & $M$ & $C$ & Type

     \\ \hline
         adult & 30162 & 13 & 2 & mixed 
         \\
         banknote & 1372 & 4 & 2 & numeric 
         \\
         breastcancer & 286 & 9 & 2 &  categorical
         \\
         car & 1728 & 6 & 4 & categorical
         \\
    COMPAS & 7214 & 6 & 2 & mixed 
    \\
    german & 1000 & 20 & 2 & mixed
    \\
    heloc(FICO) & 10459 & 23 & 2 & numeric
    \\
    ILPD & 583 & 10 & 2 &  mixed 
    \\
    liver & 345 & 6 & 2 &  numeric  
    \\
    magic & 19020 & 10 & 2 & numeric
    \\
    monks & 554 & 6 & 2 & categorical
    \\
    mushroom & 8124 & 22 & 2 & categorical
    \\
    nursery & 12959 & 8 & 5 & categorical
    \\
    tictactoe & 958 & 9 & 2 & categorical
    \\
    transfusion & 749 & 4 & 2 & numeric
    \\
    vote & 435 & 15 & 2 & categorical
    \\ \Xhline{1.2pt}
  \end{tabular}
  \label{tbl:dataset}
\end{table}

\textbf{Baselines.}
To evaluate the performance of our algorithms, three state-of-the-art rule-based interpretable classifiers are included in the performance comparison: DR-Net \cite{24qiao2021learning}, BRS \cite{2wang2017bayesian} and DRS \cite{6zhang2020diverse}. As the representatives of classic tree-based classification algorithms, classification and regression tree (CART) and random forest (RF) in the scikit-learn package \cite{pedregosa2011scikit}  are included in the comparison as well. 

\textbf{Evaluation metrics.}
In order to test the performance of our algorithms comprehensively, we choose different evaluation measures for different purposes. For predictivity, we choose the classification accuracy as the performance indicator. For interpretability, we consider the length of rule, the number of total rules and the number of distinct ``personalized" rules. 

\textbf{Parameter tuning.}
For DR-Net, we set $\lambda_1$ to be $10^{-2}$ and $\lambda_2$ to be $10^{-2}$. For BRS, we set $\alpha_{+}=\alpha_{-}=500$ and $\beta_{+}=\beta_{-}=1$. For DRS, we fix the  mode of the key hyper-parameter $\lambda$  to be Max. Through some experiments, we find that PIC will achieve better performance when $\alpha$ falls into the interval [0.7.0.9] and fPIC achieves a better performance at $\alpha =0.9, maxL=3$ .

\subsection{Performance on predictivity}
\label{sec:4.1}
\subsubsection{The dataset}

We conduct the experimental study on 16 public datasets. More precisely, 14 data sets are obtained from the UCI repository \cite{UCI}, COMPAS is a variant of the ProPublica recidivism dataset\footnote{https://www.propublica.org/datastore/dataset/compas-recidivism-risk-score-data-and-analysis} and heloc is from Fair Isaac Corporation (FICO) dataset \cite{FICO}. The detailed characteristics of these data sets are summarized in Table \ref{tbl:dataset}.  A pre-processing procedure is employed to discretize numerical values into categorical ones. That is, the equal width method is used to split the $j$th numeric feature values into $g_j$ groups, where $g_j$ is a user-specified parameter. This parameter is set to be 10 on german and adult, and it is fixed to be 5  on the other datasets. When running CART and RF, we use the original numeric futures without discretization in order to obtain better performance for these two methods.

We repeat the 5-fold  cross-validation procedure  5 times to compute the average accuracy values as the performance indicators for  predictivity. In addition, in order to finish the experiments in an acceptable time slot, $maxL$ parameter in PIC is set to be 2 on adult, 3 on heloc and mushroom, 4 on german, and 5 on nursery. For all remaining data sets, the $maxL$ parameter is fixed to be 100. Since  DR-Net and BRS can only handle binary classification problems,  we employ  the one-versus-one (OVO) strategy  to  accomplish the multi-class classification task on car and nursery. That is, we construct $C(C-1)/2$ classifiers for each pair of classes and the final predicted class label  will be determined by the voting result from the $C(C-1)/2$ classifiers.

\subsubsection{Results}
\begin{table}[tb]
  \caption{The average accuracy values based on the repeated execution of the 5-fold cross-validation procedure 5 times.}
  \centering
    \begin{adjustbox}{width=1.1\textwidth,center}

  \renewcommand\arraystretch{1.2}

  \begin{tabular}{c|ccccccc|c}
    \Xhline{1.2pt}

     \multirow{2}{*}{Dataset} & \multicolumn{2}{c}{PIC}  & \multirow{2}{*}{fPIC}  & \multirow{2}{*}{DR-Net} & \multirow{2}{*}{BRS} & \multirow{2}{*}{DRS}  & \multirow{2}{*}{CART}  & \multirow{2}{*}{RF}
     \\
      & $\alpha=0.7$ & $\alpha=0.9$ &  & & & & &

     \\ \hline
         adult & 0.754 & 0.751& 0.779 & $\pmb{0.836}$ & 0.815 & 0.459  & 0.778 & 0.821
         \\
         banknote & 0.917 & 0.971& 0.968 & 0.843 & 0.929 & 0.948 & $\pmb{0.982}$ & 0.993 
         \\
         breastcancer & 0.747 & 0.713 & 0.675 & $\pmb{0.772}$ & 0.744 & 0.746 & 0.698 & 0.758
         \\
         car &  0.701 & 0.728 & 0.754 & 0.659 & 0.661 & 0.791 & $\pmb{0.974}$ & 0.965
         \\
    COMPAS & 0.564 & 0.633 & $\pmb{0.636}$ & 0.623 & 0.623 & 0.547 & 0.630 & 0.636
    \\
    german & 0.703 & $\pmb{0.741}$ & 0.673 & 0.715 & 0.723 & 0.695 & 0.644 & 0.752
    \\
    heloc(FICO) & 0.692 & 0.690 & 0.685 & 0.693 & $\pmb{0.694}$ & 0.645 & 0.629 & 0.722
    \\
    ILPD & $\pmb{0.711}$ & 0.688 & 0.679 & 0.705 & 0.681 & 0.701 & 0.619 & 0.654
    \\
    liver & $0.584$ & 0.558 & 0.513 & 0.579 & 0.512 & 0.525 & \pmb{0.631} & 0.730
    \\
    magic & 0.672 & 0.810  & 0.801 & 0.801 &  0.784 & 0.657 & $\pmb{0.816}$ & 0.880
    \\
    monks & 0.964 & 0.972 & 0.946 & 0.747 &  0.982 & $\pmb{0.991}$ & 0.971 & 0.982
    \\
    mushroom & 0.958 & 0.989 & 0.995 &  0.995 & 0.994 & 0.993 & $\pmb{1.000}$ & 1.000
    \\
    nursery & 0.839 & 0.925 & 0.950 & 0.918 & 0.727 & 0.816 & $\pmb{0.996}$ & 0.991
    \\
    tictactoe & 0.683 & 0.971 & 0.890 & 0.404 & 0.978 & $\pmb{0.992}$ & 0.943 & 0.983
    \\
    transfusion & 0.759 & 0.762 & $\pmb{0.764}$ & 0.760 & 0.761 & 0.757 & 0.706 & 0.739
    \\
    vote & $\pmb{0.958}$ & 0.954 & 0.933 & 0.614 & 0.923 & 0.905 & 0.934 & 0.954

    \\ \Xhline{1.2pt}
  \end{tabular}
  \label{tbl:acc}
  \end{adjustbox}
\end{table}

\begin{figure}[!h]
  \centering
  \includegraphics[width=\columnwidth]{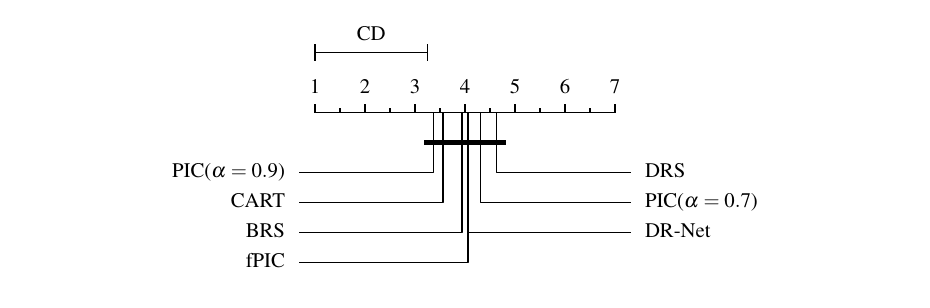}
  \caption{Bonferroni-Dunn critical difference diagrams on accuracy at a significance level of 0.05. In the figure, each method is positioned according to its average rank across all data sets. Two methods will be  connected by a thick line if their performance gap is not statistically  significant.}
  \label{fig:cd}
\end{figure}
\begin{figure}[!h]
  \centering
  \includegraphics[width=\columnwidth]{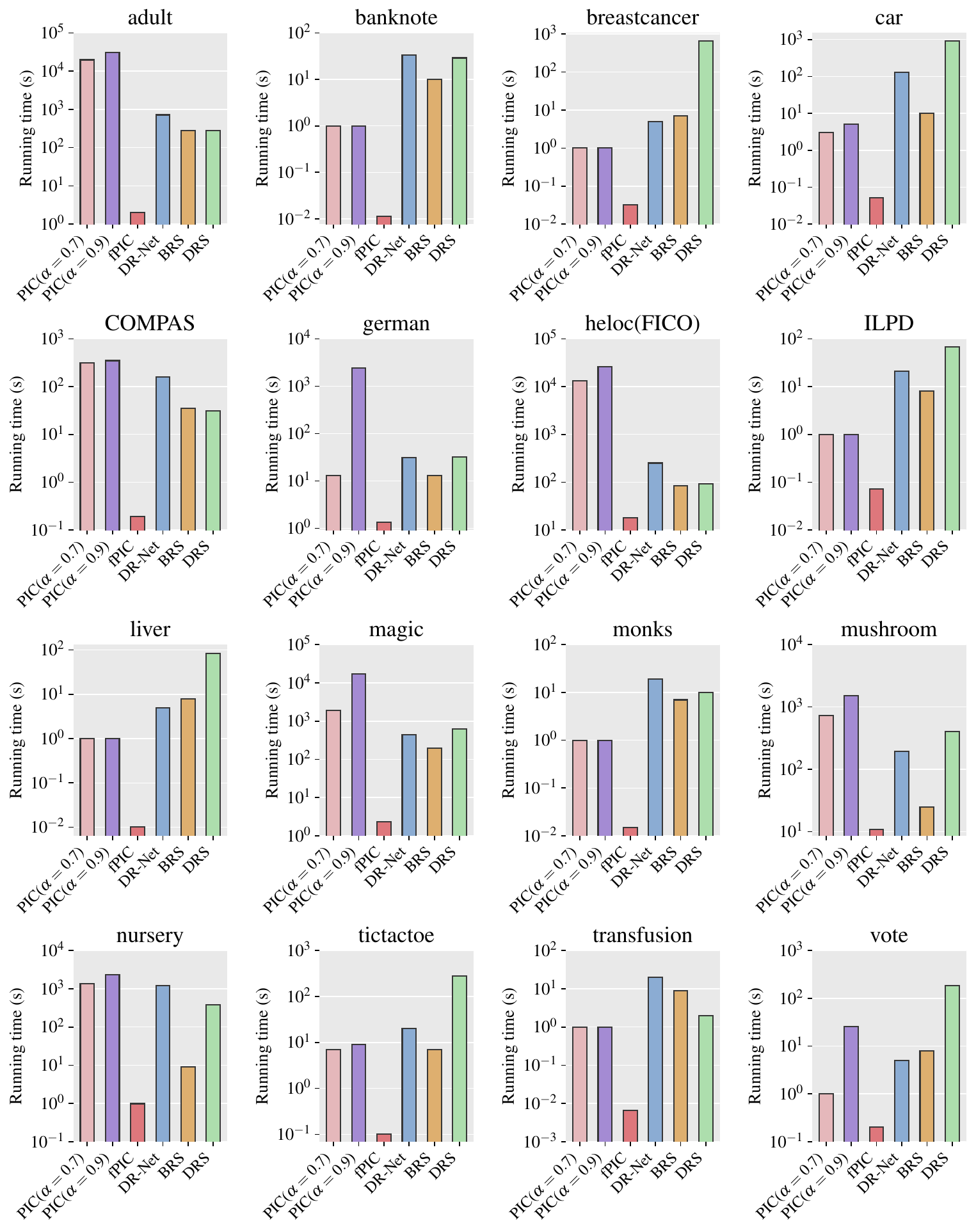}
  \caption{The average running time(s) based on the repeated execution of the 5-fold cross-validation procedure 5 times.}
  \label{fig:time}
\end{figure}

Table \ref{tbl:acc} presents the comparison result  between baseline methods and our algorithms based on the 5-fold cross-validation. The best accuracy values among five interpretable methods (ours, DR-Net, BRS, DRS and CART) on each dataset are marked in bold. The DR-Net algorithm does not perform very good on tictactoe and vote, probably because the sample size of these two datasets is  not big enough for training DR-Net. Table \ref{tbl:acc} shows that PIC and fPIC can achieve the same-level classification accuracy as those interpretable classifiers chosen in this experiment. On most of datasets, the accuracy values of our algorithms are very close to those values of the tree-based interpretable method (CART). Since fPIC is more strict on the maximal length of rules, its average accuracy is slightly lower than that of PIC. Furthermore, the performance gap between our algorithms and  RF is  within 0.02 on all datasets except adult, car and nursery. The PIC algorithm does not perform very well on these datasets probably because of the class imbalance of these datasets. 

To further check whether the performance gap among different algorithms is statistically significant, we conducted the Bonferroni-Dunn test. The significance test results ($CD=2.06$) are shown in Figure~\ref{fig:cd}. It can be observed that although PIC($\alpha=0.9$) does not significantly outperform other methods across all datasets, we can at least claim that our methods are competitive to existing state-of-the-art interpretable classification algorithms.

Figure~\ref{fig:time} displays the average running time of different classification algorithms. We can find that  PIC is very efficient on those small data sets such as banknote and liver. That is, PIC can achieve the same level efficiency as those tree-based classifiers and takes less running time than rule-based interpretable classifiers on these data sets. However,  the running time of PIC on some large datasets, especially on adult, heloc and magic, is several orders of magnitude larger than the time consumed by the others. This happens because PIC constructs a personalized model for each test sample. In essence, PIC is a lazy learning method, so it will be more time-consuming than the other eager learning methods. However, compared to other classification algorithms, including PIC, fPIC has a significant advantage in running time. This is because fPIC builds a hash table during preprocessing to record the occurrence frequencies of different itemsets in the training set, which accelerates rule evaluation and selection during prediction.

\subsection{Performance on personalization and  interpretability}
\label{sec:4.2}

To date, there is still no universally recognized precise definition of interpretability. In our case, we try to measure the performance of each algorithm in terms of interpretability via  the length of rule, the number of total rules and the comparison between the distinct ``personalized" rules and the common rules. 

\begin{table}[htb]
  \caption{The average length of rules reported by each algorithm.}
  \centering
  \renewcommand\arraystretch{1.2}
  \begin{tabular}{c|cccccc}
    \Xhline{1.2pt}
     \multirow{2}{*}{Dataset} & \multicolumn{2}{c}{PIC} & \multirow{2}{*}{fPIC} & \multirow{2}{*}{DR-Net} & \multirow{2}{*}{BRS} & \multirow{2}{*}{DRS}
     \\
     & $\alpha=0.7$ & $\alpha=0.9$& & & & 

     \\ \hline
         adult & 1.91 & 2.00 & 2.71 &  10.50 &  2.95 &  9.55 
         \\
         banknote & 1.44 & 2.11 & 1.78 & 9.67 & 2.97 & 2.59
         \\
         breastcancer & 1.62 & 2.86 & 2.02 & 38.00 & 2.98 & 4.03 
         \\
         car & 1.08 & 1.38 & 2.56 & 16.15 & 1.45 & 4.82 
         \\
    COMPAS & 2.96 & 3.73 & 2.11 & 53.80 & 3.00 & 3.21  
    \\
    german & 1.39 & 3.77 & 2.29 & 24.18 & 3.00 & 11.59
    \\
    heloc(FICO) &1.75 & 2.84  & 2.89 & 23.85 &  2.90 &  10.01 
    \\
    ILPD  & 1.00 & 1.89 & 1.96 & 43.62 & 2.82 & 5.70
    \\
    liver  & 1.22 & 2.51 & 1.92 & 30.00 & 3.00 & 3.60 
    \\
    magic & 1.53  & 3.83  & 2.59 & 8.99  & 3.00  & 3.62 
    \\
    monks & 1.57 & 1.95 & 2.00 & 9.81 & 3.00 & 2.41
    \\
    mushroom & 2.07 & 2.91 & 1.59 & 11.82 & 2.84 & 6.52
    \\
    nursery & 2.07 & 2.91 & 2.79 & 8.61 & 1.41 & 2.72 
    \\
    tictactoe & 1.05 & 2.85 & 2.46 & 13.86 & 3.00 & 3.92 
    \\
    transfusion & 1.00 & 1.17 & 1.30 & NaN & 3.00 & 2.45 
    \\
    vote & 1.13 & 2.04 & 1.42 & 45.00 & 2.97 & 6.33 

    \\ \Xhline{1.2pt}
  \end{tabular}
  \label{tbl:ruleL}
\end{table}

Table \ref{tbl:ruleL} shows the average length of rules reported by four methods. For the our algorithms, we calculate the average length of all the rules found in a 5-fold cross-validation. In comparison, we use the average length of the 5 rule sets reported by the other methods. The rules reported by DR-Net are  longer than the ones reported by other methods on most of the datasets except transfusion. DR-Net does not find any rules on the transfusion data set. In contrast, the rules found by our algorithms are shorter and their lengths  are all less than $4$. The rules found by PIC($\alpha=0.7$) are shorter than those found by PIC($\alpha=0.9$) and fPIC.

\begin{table}[htb]
  \caption{The average number of rules reported by each algorithm.}
  \centering
  \renewcommand\arraystretch{1.2}
  \begin{tabular}{c|cccccc}
    \Xhline{1.2pt}

     \multirow{2}{*}{Dataset} & \multicolumn{2}{c}{PIC}  & \multirow{2}{*}{fPIC} & \multirow{2}{*}{DR-Net} & \multirow{2}{*}{BRS} & \multirow{2}{*}{DRS}
     \\
     & $\alpha=0.7$ &  $\alpha=0.9$  & & & & 

     \\ \hline
        adult & 5.0 & 381.4 & 828.6 & 10.0 & 3.8 & 6.6 \\
        banknote & 7.0 & 32.8 & 43.6 & 48.4 & 7.6 & 19.8 \\
        breastcancer & 20.2 & 30.4 & 37.0 & 50.0 & 10.0 & 63.6 \\
        car & 13.4 & 41.6 & 75.4 & 283.0 & 66.6 & 70.4 \\
        COMPAS & 34.8 & 85.8 & 88.4 & 21.6 & 3.2 & 4.6 \\
        german & 4.8 & 111.6 & 129.2 & 22.2 & 5.8 & 5.6 \\
        heloc (FICO) & 126.6 & 624.6 & 1223.2 & 19.4 & 4.2 & 3.2 \\
        ILPD & 2.2 & 40.8 & 61.6 & 11.0 & 8.0 & 16.4 \\
        liver & 12.0 & 29.8 & 35.8 & 50.0 & 5.8 & 23.0 \\
        magic & 22.2 & 371.6 & 534.8 & 21.0 & 3.2 & 17.2 \\
        monks & 11.4 & 20.6 & 49.8 & 50.0 & 7.4 & 13.4 \\
        mushroom & 10.6 & 21.6 & 100.0 & 49.2 & 8.2 & 19.8 \\
        nursery & 147.2 & 328.8 & 424.0 & 258.0 & 75.6 & 16.0 \\
        tictactoe & 10.0 & 21.6 & 83.8 & 48.8 & 8.0 & 28.4 \\
        transfusion & 3.0 & 11.2 & 11.0 & NaN & 1.2 & 4.0 \\
        vote & 4.4 & 10.8 & 29.0 & 50.0 & 7.2 & 21.2 
    \\ \Xhline{1.2pt}
  \end{tabular}
  \label{tbl:ruleN}
\end{table}
\begin{figure}[!h]
  \centering
  \includegraphics[width=0.95\columnwidth]{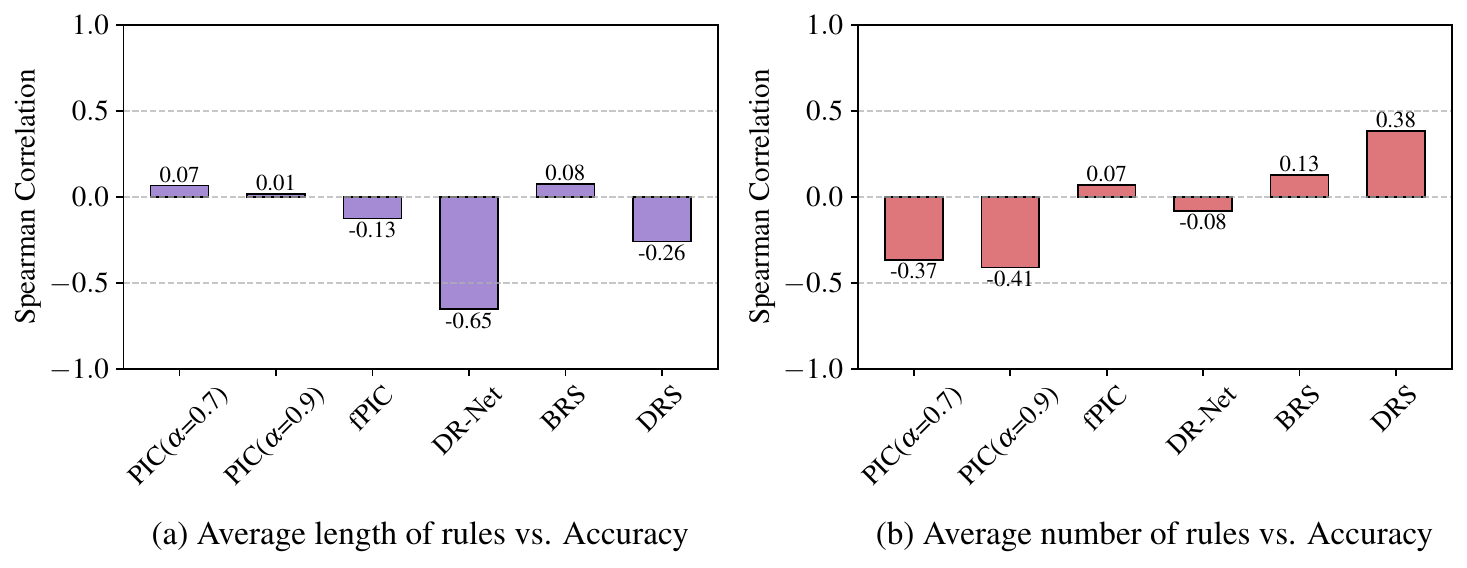}
  \caption{Spearman correlation coefficients between classification accuracy and both the average length of rules and the average number of rules.}
  \label{fig:correlation}
\end{figure}

\begin{figure}[!h]
  \centering
  \includegraphics[width=0.9\columnwidth]{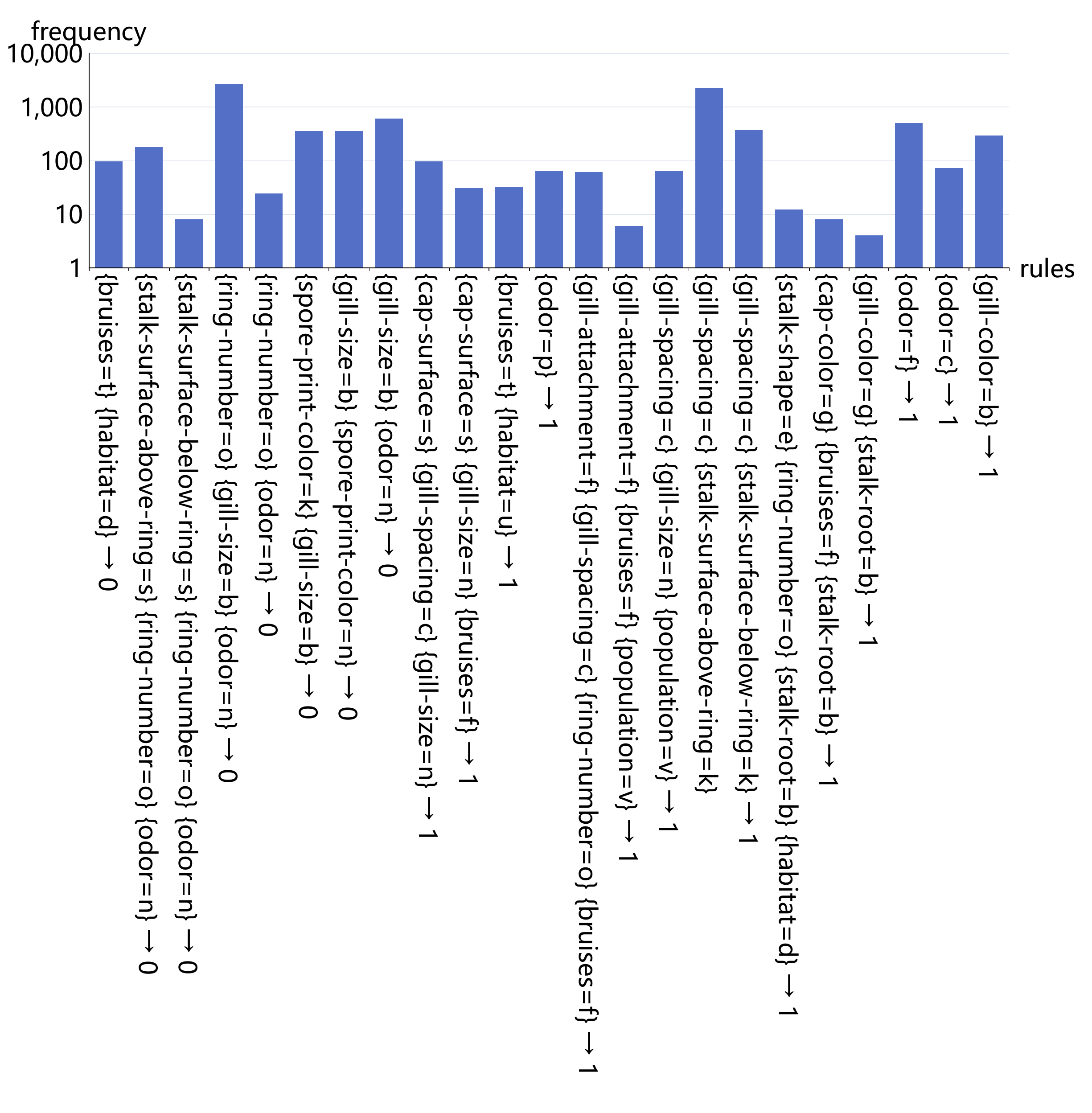}
  \caption{The distribution of the rules reported by PIC on mushroom in a 5-fold cross-validation when $\alpha=0.9$.}
  \label{fig:mushroom}
\end{figure}

\begin{table}[!h]
  \caption{The rules reported by PIC on mushroom in a 5-fold cross-validation when $\alpha=0.9$.}
  \renewcommand\arraystretch{1.2}
  \centering
  \footnotesize
  \begin{adjustbox}{width=1.1\textwidth,center}
  \begin{tabular*}{1.1\textwidth}{l|l}
    \Xhline{1.2pt}
    \multicolumn{2}{c}{Common rules}
    \\ \hline
    Rules & Method
     \\ \hline
          \{bruises=t\} \{habitat=d\} $\rightarrow$ 0 & DRS
        \\
     \{stalk-surface-above-ring=s\} \{ring-number=o\} \{odor=n\} $\rightarrow$ 0 & DRS
    \\
    \{stalk-surface-below-ring=s\} \{ring-number=o\} \{odor=n\} $\rightarrow$ 0 & DRS
    \\
    \{ring-number=o\} \{gill-size=b\} \{odor=n\} $\rightarrow$ 0 & DRS
    \\
    \{ring-number=o\} \{odor=n\} $\rightarrow$ 0  & DRS
    \\
    \{gill-size=b\} \{odor=n\} $\rightarrow$ 0  & DRS
    \\
    \{bruises=t\} \{habitat=u\} $\rightarrow$ 1 & DR-Net
    \\
    \{odor=p\} $\rightarrow$ 1  & BRS, DRS
    \\
    \{gill-attachment=f\} \{gill-spacing=c\} \{ring-number=o\} \{bruises=f\} $\rightarrow$  1 & DRS
    \\
    \{gill-attachment=f\} \{bruises=f\} \{population=v\} $\rightarrow$ 1  & DRS
    \\
    \{gill-spacing=c\} \{gill-size=n\} \{population=v\} $\rightarrow$  1 & BRS, DRS \\
    \{gill-spacing=c\} \{stalk-surface-above-ring=k\} $\rightarrow$  1 & DR-Net, BRS, DRS\\
    \{gill-spacing=c\} \{stalk-surface-below-ring=k\} $\rightarrow$  1 & DRS \\
    \{stalk-shape=e\} \{ring-number=o\} \{stalk-root=b\} \{habitat=d\} $\rightarrow$  1 & DRS \\
    \{cap-color=g\} \{bruises=f\} \{stalk-root=b\} $\rightarrow$ 1 & DRS \\
    \{gill-color=g\} \{stalk-root=b\} $\rightarrow$ 1  & BRS \\
    \{odor=f\} $\rightarrow$  1  & DR-Net, BRS, DRS \\
    \{odor=c\} $\rightarrow$  1 & DRS \\
    \{gill-color=b\} $\rightarrow$  1 & DR-Net, BRS, DRS \\

    \hline
    \multicolumn{2}{c}{``Personalized" rules}
    \\ \hline
    \multicolumn{2}{l}{\{spore-print-color=k\} \{gill-size=b\} $\rightarrow$ 0 } \\
    \multicolumn{2}{l}{\{gill-size=b\} \{spore-print-color=n\} $\rightarrow$ 0 } \\
    \multicolumn{2}{l}{\{cap-surface=s\} \{gill-spacing=c\} \{gill-size=n\} $\rightarrow$ 1 } \\
    \multicolumn{2}{l}{\{cap-surface=s\} \{gill-size=n\} \{bruises=f\} $\rightarrow$ 1 }

    \\ \Xhline{1.2pt}
  \end{tabular*}
  \end{adjustbox}
  \label{tbl:ruleP}
\end{table}

Table \ref{tbl:ruleN} shows the average number of rules reported by four methods. The PIC and fPIC algorithm can find more rules than any other algorithms on COMPAS, heloc, magic and transfusion, which means that our method is more likely to find more ``personalized" rules for distinct samples. Since fPIC does not include a pruning strategy, it discovers even more rules than PIC.

To evaluate the impact of different rule characteristics on the predictive performance of the model, we compute the Spearman correlation coefficients between accuracy and both the average length of the rules and the average number of rules for different methods. As shown in Figure~\ref{fig:correlation}(a), our algorithms effectively control the length of the rules, which does not result in a significant correlation with accuracy. In contrast, for other eager learning methods, excessively long rules may lead to overfitting, thereby reducing accuracy and exhibiting a negative correlation. Figure~\ref{fig:correlation}(b) indicates that for our algorithms, a larger number of rules suggests fewer shared rules, making the data more difficult to classify, leading to a negative correlation. However, for other eager learning methods, an increase in the number of rules generally improves accuracy, showing a positive correlation.

Figure~\ref{fig:mushroom} shows the distribution of the rules reported by PIC on mushroom in a 5-fold cross-validation  when $\alpha=0.9$. PIC reports 23 rules, 8 of them are from class 0 and 15 of them are from class 1.  We can find that there are 2 rules, (\{ring-number=o\} \{gill-size=b\} \{odor=n\} $\rightarrow$ 0) and (\{gill-spacing=c\} \{stalk-surface-above-ring=k\} $\rightarrow$ 1), which appear more times than others. And the other rules only appear less than 1000 times in the experiment.

We consider a rule $r_1=(s_1,y_1)$ reported by PIC is a common rule when there is a rule $r_2=(s_2,y_2)$ reported by other methods that satisfies $s_1\subseteq s_2\land y_1=y_2$. Table \ref{tbl:ruleP} shows the common rules found on mushroom by PIC and other methods and the ``personalized" rules only reported by PIC when $\alpha$ is set to 0.9. We find 4 ``personalized" rules in this experiment, 2 for class 0 and 2 for class 1. Their frequency values are all  in the range of 10 to 1000, so they do not appear by accident. This fact demonstrates  that the PIC algorithm really can find some ``personalized" rules which cannot be discovered by other methods.
\subsection{Parameter sensitivity}
\label{sec:4.3}

\begin{figure}[t]
  \centering
  \includegraphics[width=\columnwidth]{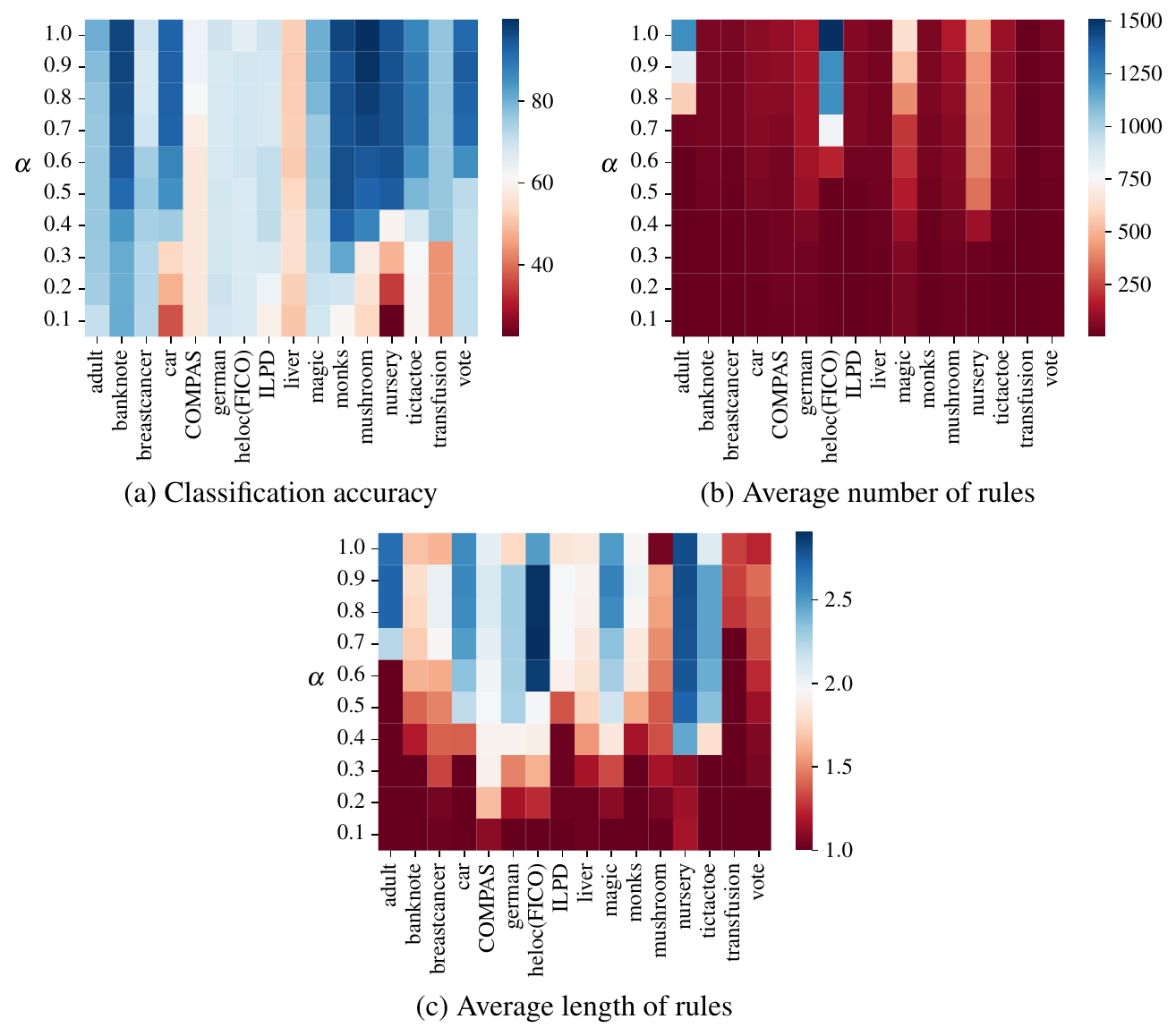}
  \caption{Impact of varying $\alpha$ on classification accuracy, average number of rules, and average length of rules.}
  \label{fig:alpha}
\end{figure}

Both PIC and fPIC are designed to identify the local optimal matching rule for each sample, following the same underlying principles and rule evaluation function. Consequently, the weighting factor $\alpha$ in Equation~(\ref{eq:A}) and the maximum itemset length $maxL$ have a largely similar impact on performance of both PIC and fPIC. To enhance experimental efficiency, we conduct the parameter sensitivity analysis solely on fPIC and present the results as follows:

The impact of the weighting factor $\alpha$ on fPIC's classification accuracy, average number of rules, and average length of rules is shown in Figure~\ref{fig:alpha}.

\begin{itemize}
    \item \textbf{Classification accuracy}: As $\alpha$ increases, the model's predictive performance improves. However, when $\alpha = 1.0$, accuracy slightly decreases on some datasets. This suggests that the linear combination of precision and recall in Equation~(\ref{eq:A}) is critical for rule evaluation.
    \item \textbf{Average number of rules}: In most datasets, the number of rules remains stable as $\alpha$ increases. In a few cases, the number of rules increases, indicating that the model identifies more discriminative rules, reducing the number of shared rules.
    \item \textbf{Average length of rules}: The average rule length follows a similar trend to $\alpha$. When precision has a higher weight in the rule evaluation function, the model tends to generate longer rules.
\end{itemize}

\begin{figure}[t]
  \centering
  \includegraphics[width=\columnwidth]{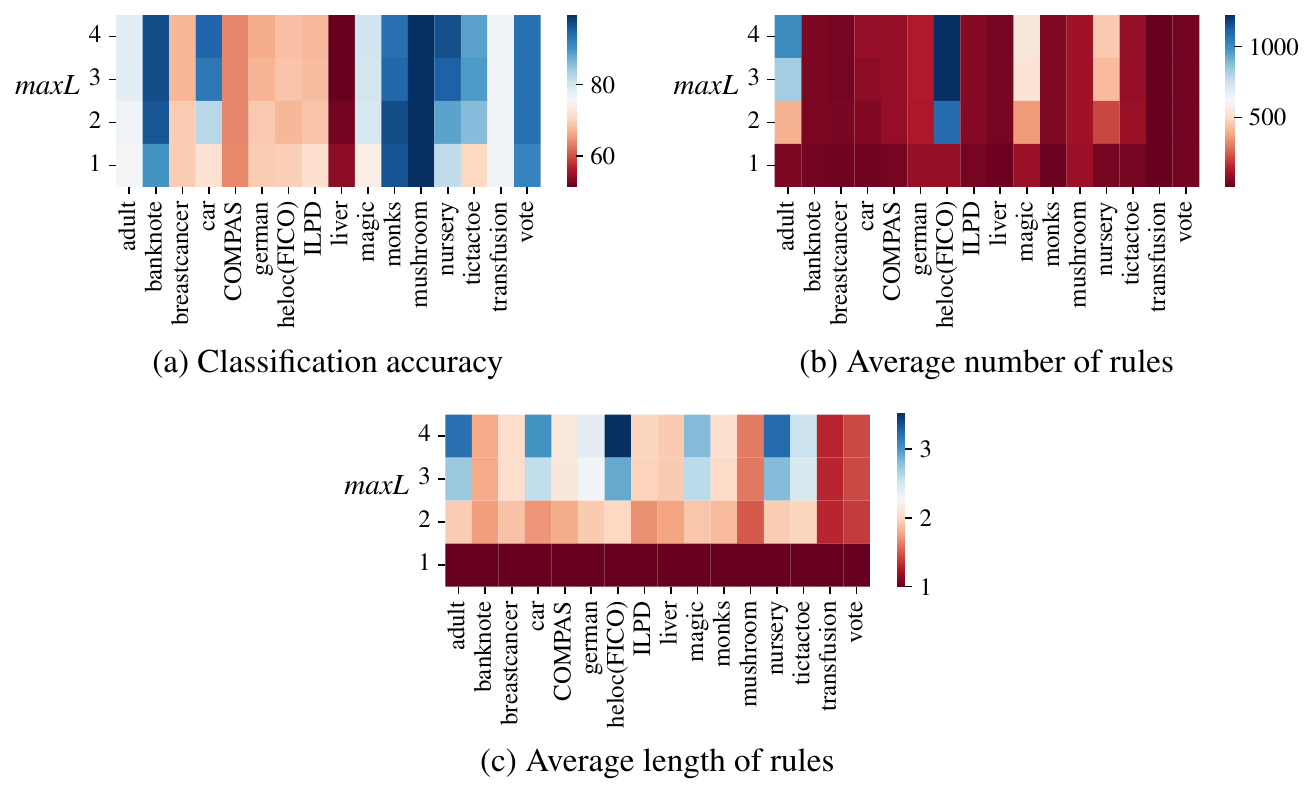}
  \caption{Impact of varying $maxL$ on classification accuracy, average number of rules, and average length of rules.}
  \label{fig:len}
\end{figure}

The impact of the maximum itemset length $maxL$ on fPIC's classification accuracy, average number of rules, and average number of rules is shown in Figure~\ref{fig:len} ($maxL$ is ranged from 1 to 4 because using values larger than 4 will exceed the memory limit of our computer):

\begin{itemize}
    \item \textbf{Classification accuracy}: When $maxL$ reaches 3, its impact on classification accuracy becomes minimal, while the predictive performance of model remains good.
    \item \textbf{Average number of rules}: $maxL$ has little effect on the number of rules for most datasets. In a few datasets, the number of rules increases with the increase of $maxL$, indicating that the model identifies more discriminative rules, reducing the number of shared rules.
    \item \textbf{Average rule length}: When $maxL$ exceeds 1, the average rule length increases with the increase of  $maxL$ for most datasets. However, shorter rule lengths provide better interpretability, so $maxL$ should be controlled as much as possible while maintaining model accuracy.
\end{itemize}

\subsection{A real application scenario}

\subsubsection{The dataset}
  
To illustrate why personalized and interpretable classification methods should be developed, here we use the application of predicting breast cancer metastasis \cite{14jahid2014personalized} as a real example. As a complex and heterogeneous disease,  breast cancer typically has many molecular subtypes. Hence, classifiers constructed for one cohort  often cannot achieve good performance on other cohorts. To tackle this issue, one feasible solution is to assume that each patient belongs to a distinct subtype and construct different classification models for different patients \cite{14jahid2014personalized}. 

The breast cancer metastasis dataset is derived from \cite{14jahid2014personalized}, which is composed of 1522 features and 581 samples. In our experiments, we choose 30 samples as the test set and the remaining samples are used as the training set. That is, there are 551 samples in the training set and 30 samples in the test set. The class labels are binary so both DR-Net and BRS can handle this data set  as well. All the 1522 features are numerical, so we employ the equal width method with $g_j=3$ to transform the $j$th numeric feature values into categorical ones in our methods. 

\subsubsection{Results}
\begin{table}[t]
  \caption{Experimental results on the breast cancer dataset.}
  \centering
  \renewcommand\arraystretch{1.2}
  \footnotesize
  \begin{tabular}{c|cccc}
    \Xhline{1.2pt}

     Methods & Accuracy & Time (s) & \#Rule & Length

     \\ \hline
    PIC($\alpha=0.7$) & 0.633 & 64188 & 2 & 2.00 \\
    PIC($\alpha=0.9$) & 0.633 & 84999 & 26 & 2.00 \\
    fPIC & 0.700 & 1 & 29 & 1.00 \\
    DR-Net & 0.366 & 43 & NaN & NaN \\
    BRS & 0.533 & 37749 & 38 & 2.76 \\
    DRS & 0.366 & 192 & NaN & NaN \\
    CART & 0.500 & 1  & - & - \\
    RF & 0.566 & 1 & - & -

    \\ \Xhline{1.2pt}
  \end{tabular}
  \label{tbl:gene}
\end{table}
We conduct the following experiments on a workstation with an AMD Ryzen 5 5600X 6-Core Processor(3.70 GHz) and 32GB memory, which runs approximately twice as fast as the workstation used in \ref{sec:4.1} and \ref{sec:4.2}. When running DRS, we encounter an error when transforming  the type of variables from type(`O') to type(`float64') in the digitize function from numpy package. We have written a function to replace it to continue the experiment. As $maxL$ increases, the rule search space of our algorithms expands, reducing its running efficiency. To finish the experiment within 24 hours, $maxL$ is set to be 2 in the PIC algorithm. For the parameter $\alpha$ in PIC, we still set it to 0.7 and 0.9 for comparison. Due to memory constraints, when  $maxL$  is greater than 1, fPIC fails to execute during the preprocessing stage while constructing the hash table. Hence, in fPIC,  $maxL$  is set to 1 and  $\alpha$  to 0.9. Besides, all other methods parameters are set the same as what those in \ref{sec:4.1} and \ref{sec:4.2}. 

The experimental results are shown in Table \ref{tbl:gene}. DR-Net and DRS accomplish the classification task very fast but they do not find any rules. BRS finds a rule set of 38 rules, 9 of them have a length of 2 and 29 of them have a length of 3. When $\alpha$ is set to be 0.7, PIC can find 2 different  rules. And when $\alpha$ is set to be 0.9, PIC can find 26 distinct rules. fPIC reported 29 distinct rules of length 1, which is nearly the same as the number of test samples. This is because  $maxL$  is set to 1, leading to the discovery of more ``personalized" rules. All the rules reported by PIC have a length of 2, which means that better rules might be found if we further increase  the $maxL$ parameter.  The results in Table 11  also show that PIC can find more ``personalized" rules when $\alpha$ is set to be 0.9 at the cost of  consuming more running time. More importaly, our algorithms can achieve better predictive accuracy than all other competing classification methods in the performance comparison. It demonstrates the rationale of developing personalized interpretable classifiers in real applications such as cancer metastasis prediction.

\section{Conclusion}
In this paper, we introduce the personalized interpretable classification issue and present two algorithms: a greedy algorithm called PIC and a fast rule discovery algorithm called fPIC. Both are designed to identify a personalized rule for each individual test sample. To demonstrate the effectiveness of our algorithms, we conduct a series of experiments on some real  data sets. The experimental results show that such personalized interpretable classifiers  can achieve good performance both on predictivity and interpretability.

Overall, we formally introduce a new data mining problem, namely  personalized interpretable classification. By solving this new classification issue, we can identify some ``personalized" rules that cannot be found by existing interpretable classification methods. PIC achieves comparable predictive accuracy but has a longer runtime when the dataset is large. fPIC can quickly discover personalized rules, but the number of shared rules found is smaller, and the classification accuracy is slightly lower. Additionally, when the maximal itemset length is large, it may exceed memory limits. Both PIC and fPIC have their strengths and weaknesses, and the appropriate classification algorithm should be chosen based on the specific situation. In the future, we will further develop more effective algorithms for mining personalized rules from large data sets in different application domains.

\section*{Acknowledgments}

This work has been supported by the Natural Science Foundation of China
iunder Grant No. 62472064.




\bibliographystyle{elsarticle-num} 
\bibliography{PIC}





\end{document}